\documentclass[letterpaper, 10 pt, conference]{ieeeconf}  
\IEEEoverridecommandlockouts 
\usepackage{lipsum}
\usepackage{bm}
\usepackage{amsmath}
\usepackage{amssymb}
\usepackage{tikz}
\usepackage{xcolor}
\usetikzlibrary{shapes.geometric, arrows.meta, patterns.meta}
\usepackage{cleveref}
\usepackage{siunitx}

\newcommand{\bb}[1]{\bm{#1}}								
\newcommand{\id}[3]{{}_\mathit{#1}{#2}_\mathit{#3}}						
\newcommand{\iddot}[3]{\id{#1}{\dot{#2}}{#3}}		

\newcommand{\idbf}[3]{\id{#1}{\bb{#2}}{#3}}					
\newcommand{\idbfdot}[3]{\id{#1}{\dot{\bb{#2}}}{#3}}		
\newcommand{\idbfhat}[3]{\id{#1}{\hat{\bb{#2}}}{#3}}		

\newcommand{\func}[2]{#1\!\left(#2\right)}
\newcommand{\M}{\mathcal{M}}
\newcommand{\U}{\mathcal{U}}
\newcommand{\TpM}{\bb{T}_\mathit{p}\M}
\newcommand{\RI}{\mathbb{R}}
\newcommand{\RII}{\mathbb{R}^2}
\newcommand{\RIII}{\mathbb{R}^3}
\newcommand{\SOII}{SO(2)}

\newcommand{\pfpx}[2]{\frac{\partial#1}{\partial#2}}		

\definecolor{col_red}{HTML}{A8515D}
\definecolor{col_orange}{HTML}{F3B85F}
\definecolor{col_beige}{HTML}{FCECCA}
\definecolor{col_grey}{HTML}{E6E6E6}
\definecolor{col_sage}{HTML}{B4D6C5}
\definecolor{col_teal}{HTML}{5FA9B4}

\title{\LARGE \bf
Consistent Pose Estimation of Unmanned Ground Vehicles through Terrain-Aided Multi-Sensor Fusion on Geometric Manifolds
}

\author{Alexander Raab$^{1}$, Stephan Weiss$^{2}$, Alessandro Fornasier$^{2}$, Christian Brommer$^{2}$ and Abdalrahman Ibrahim$^{1}$%
\thanks{*This work was supported by the FFG project Salto, funded under the "ICT of the Future" program by the Austrian Federal Ministry for Climate Action, Environment, Energy, Mobility, Innovation, and Technology (BMK).}%
\thanks{*This work has received funding from the Army Research Office under Cooperative Agreement Number W911NF-21-2-0245.}
\thanks{$^{1}$Alexander Raab and Abdalrahman Ibrahim are with AGILOX Services GmbH, Austria {\tt\small \{firstname.lastname\}@agilox.net}}%
\thanks{$^{2}$Stephan Weiss, Alessandro Fornasier and Christian Brommer are with the Control
	of Networked Systems Group, University of Klagenfurt, Austria 
        {\tt\small \{firstname.lastname\}@ieee.org}}%
\thanks{\textbf{Preprint version, accepted Jun/2025 (IROS), DOI to follow~\copyright IEEE.}}
}

\begin{document}

\maketitle
\thispagestyle{empty}
\pagestyle{empty}

\begin{abstract}
Aiming to enhance the consistency and thus long-term accuracy of Extended Kalman Filters for terrestrial vehicle localization, this paper introduces the \textit{Manifold Error State Extended Kalman Filter} (M-ESEKF). By representing the robot's pose in a space with reduced dimensionality, the approach ensures feasible estimates on generic smooth surfaces, without introducing artificial constraints or simplifications that may degrade a filter's performance. The accompanying measurement models are compatible with common loosely- and tightly-coupled sensor modalities and also implicitly account for the ground geometry. We extend the formulation by introducing a novel correction scheme that embeds additional domain knowledge into the sensor data, giving more accurate uncertainty approximations and further enhancing filter consistency. The proposed estimator is seamlessly integrated into a validated modular state estimation framework, demonstrating compatibility with existing implementations. Extensive Monte Carlo simulations across diverse scenarios and dynamic sensor configurations show that the M-ESEKF outperforms classical filter formulations in terms of consistency and stability. Moreover, it eliminates the need for scenario-specific parameter tuning, enabling its application in a variety of real-world settings. 

\end{abstract}

\section{INTRODUCTION}
With advances in technology and research, unmanned ground vehicles (UGVs) have become increasingly relevant for applications across diverse domains, including agriculture \cite{olivera2021} and logistics \cite{fragapane2022}. 
Accurate and consistent localization is generally indispensable for these robots, as it directly impacts key capabilities like autonomous navigation and control. Despite the development of more sophisticated techniques, variations of the Extended Kalman Filter (EKF) remain widely used for this task \cite{urrea2021}. The surface-bound nature of terrestrial vehicles however introduces unique challenges. Even though extensive research regarding the pose estimation of UGVs has been conducted, many authors do not explicitly leverage this property. They either estimate unrestricted 3D poses \cite{zhang2018} or implicitly assume flat environments with planar formulations \cite{teslic2011}. To better utilize prior knowledge and to avoid model simplifications, other classical EKF-based techniques tend to impose constraints on their results \cite{zheng2018}. Alternatively, some strategies incorporate domain knowledge through virtual measurements, for instance via artificial landmarks \cite{trevor2010} or pseudo-measurements obtained from topological information \cite{klein2010}, to refine their predictions. Such constraints can however negatively impact a filter's confidence, thus degrading its long-term accuracy. Addressing these issues, recent publications have adopted manifold-based estimators, which demonstrate improvements in average consistency by inherently guaranteeing feasible pose estimates along the surface \cite{brossard2020}. These approaches have been effectively utilized in localization tasks, such as magnetic inspection robots navigating smooth metal structures, employing a Particle Filter \cite{chahine2022} or an Invariant Extended Kalman Filter \cite{starbuck2021}. Others leverage motion manifolds for odometry-based dead reckoning on curved surfaces \cite{zhang2021}.

As demonstrated in these publications, the drivable terrain in many UGV applications can be modeled as smooth surfaces embedded in 3-dimensional Euclidean space. Analytically, these represent differentiable Riemannian 2-manifolds. Thus, each point on the surface can be mapped to a 2-dimensional Euclidean chart via a suitable chart map, a continuous and invertible bijection. A limitation to differentiable surfaces furthermore allows for defining a basis of the tangent space through directional derivatives at each point.

Inspired by the manifold-based methods, this paper introduces a \textit{Manifold Error State Extended Kalman Filter} (M-ESEKF) as an extension to classical EKFs for UGV pose estimation. We propose a propagation strategy directly on the chart as well as measurement models that inherently account for surface geometry to improve overall consistency without manual tuning. Additionally, we outline a novel approach that also considers terrain information within the measurement uncertainties, further stabilizing the filter's confidence.

Treating the vehicle as a planar robot moving along a curved manifold, the position is tracked on the chart, while the orientation is estimated in the tangent space. Hence, the state estimation problem simplifies from the full 3D pose on $SE(3)$ to the product space $\mathbb{R}^2 \times SO(2)$ with only three degrees of freedom. This guarantees feasible pose estimates without artificial constraints. The assumption of local planarity is inspired by the 2-dimensional odometry models commonly utilized for state propagation in wheeled and tracked vehicles \cite{thrun2005}. Based on measurements of wheel encoders, these estimate a robot's motion within the current tangent plane. To mitigate drift inherent in relative positioning and to enhance overall precision, this data is combined with global reference measurements, such as loosely-coupled pose outputs of map-based localization systems \cite{chalvatzaras2023}, enabling seamless integration into existing platforms. To improve robustness and expand potential use-cases, infrastructure-based range measurements, such as those from ultra-wideband (UWB) tags \cite{zafari2019}, are also incorporated as complementary corrections in a tightly-coupled fashion. Like the state representation, the proposed measurement models also inherently incorporate the manifold geometry contributing to consistent results. These formulations are however defined in 3D space, closely reflecting the physical interpretation of sensor outputs and possibly misrepresenting the true uncertainties confined to the surface. Thus, we outline a novel strategy for projecting measurements and their covariances onto the manifold to process the corrections directly within the chart. This allows for a more accurate uncertainty representation along the manifold, further enhancing estimator consistency.  Finally an Error State Extended Kalman Filter (ESEKF) is employed for data fusion due to the improved linearization behavior and long-term accuracy for pose tracking tasks compared to classical EKFs \cite{roumeliotis1999}. 

While the proposed filter is applicable to generic smooth surfaces, bivariate b-spline models were selected for verification due to their versatility in approximating diverse terrain geometries. To ensure consistency in data processing and enabling a direct comparison, both the classical constrained approach and the newly developed filter were integrated into a validated ESEKF-based framework for modular state estimation, \textit{MaRS} \cite{brommer2021}. Simulated experiments with dynamically changing sensor modalities demonstrate the assumed performance improvements, verifying the feasibility of our method and illustrating its potential for broader applications.

\section{METHODOLOGY}
In order to detail the methods adopted for the proposed estimator, this chapter introduces the theoretical concepts necessary for its manifold-based formulation. These fundamentals are then employed to discuss the selected state representation, the propagation model and the sensor corrections utilized within the M-ESEKF. 

\subsection{Differentiable Manifolds}
An $n^\text{th}$-dimensional manifold $\M$ is a topological space $\left(\M, \Theta\right)$, where each point $\idbf{}{p}{}$ has a neighborhood that is a homeomorphism to the $\mathbb{R}^n$. If $\forall\idbf{}{p}{} \in \M$ exists a subspace $\U \in \Theta$ and a function $\sigma:\U \rightarrow \func{\sigma}{\U} \in \mathbb{R}^n$ so that
\begin{align}
	\sigma &\text{ is invertible, thus } \exists\sigma^{-1}:\func{\sigma}{\U} \rightarrow\U \\
	\sigma &\text{ is continuous} \\ 
	\sigma^{-1} &\text{ is continuous,}
\end{align}
then the pair $\left(\U, \sigma\right)$ is regarded as a chart and the $\sigma$ as a chart map. The methodology employed in this paper focuses on the subclass of differentiable manifolds, allowing the definition of a tangent space $\TpM$ at each point $\bb{p} \in \M$ as the set of all possible tangent vectors of smooth curves on $\M$ passing through $\bb{p}$. Further specifying this categorization, the surfaces discussed here are considered parallelizable manifolds. For these, a smooth vector field $\{\idbf{}{B}{1},\dots,\idbf{}{B}{n}\}$ exists, such that the tangent vectors $\{\idbf{}{B}{1}(\bb{p}),\dots,\idbf{}{B}{n}(\bb{p})\}$ form a basis of the tangent space $\TpM$ at each point $\bb{p} \in \M$. More rigorous definitions and additional information on these concepts are discussed in literature \cite{tu2011}.

\begin{figure}[t]
	\centering
	\begin{tikzpicture}[scale=.85]
	
	\clip (-3.3,-1.17) rectangle (6,6.9);

	\node (M) at (0,0) {};
	\begin{scope}[scale=.7, rotate around={0:(M)}, shift={(M)}]
		\node (M1) at (0,0) {};
		\node (M2) at (-4,6) {};
		\node (M3) at (1.5,0) {};
		\node (M4) at (-2.5,6) {};
		\node (M5) at (4.5,2.5) {};
		\node (M6) at (0.5,8.5) {};
		\node (M7) at (8.5,2) {};
		\node (M8) at (4.5,8) {};
		\draw  plot[smooth, tension=.7] coordinates {(0,0) (1.5,0) (4.5,2.5) (8.5,2) };
		\begin{scope}[scale=1, shift={(-2,3)}]
			\draw[very thin]  plot[smooth, tension=.7] coordinates {(0,0) (1.5,0) (4.5,2.5) (8.5,2)};
		\end{scope}
		\begin{scope}[scale=1, shift={(-4,6)}]
			\draw  plot[smooth, tension=.7] coordinates {(0,0) (1.5,0) (4.5,2.5) (8.5,2) };
		\end{scope}
		\draw (M2.center) edge (M1.center);
		\draw[very thin]  (M4.center) edge (M3.center);
		\draw[very thin]  (M6.center) edge (M5.center);
		\draw  (M8.center) edge (M7.center);
	\end{scope}
	
	\node[inner sep=.1] (T0) at (1.1,3.5) {};
	\begin{scope}[scale=1, rotate around={0:(T0)}, shift={(T0)}]
		
		\begin{scope}[scale=1.5]

			\node (T1) at (0.75,-1.2) {};
			\node (T2) at (-0.75,1.2) {};
			\node (T3) at (1.2,.9) {};
			\node (T4) at (-1.2,-.9) {};
			\node (T5) at  (1.95,-.3) {};
			\node (T6) at (.45,2.1) {};
			\node (T7) at  (-1.95,.3)  {};
			\node (T8) at (-.45,-2.1) {};
			
			\draw[dashed, fill=col_teal!25!white ] (T5.center) -- (T6.center) -- (T7.center) -- (T8.center) --cycle;
			\draw[dashed, very thin]  (T1.center) edge (T2.center);
			\draw[dashed, very thin]  (T3.center) edge (T4.center);
		\end{scope}
		
		\draw[thick, -latex]  (T0.center) -- (2.75,2.05);
		\draw[thick, -latex]  (T0.center) -- (-1.72, 2.75);
		
		\begin{scope}[scale=1]
			\node (R1) at (1.45,0.05) {};
			\node (R2) at (-0.95,1.25) {};
			\node (R3) at (-1.45,-0.05) {};
			\node (R4) at (0.9,-1.25) {};
			\node (R5) at (.95,0.4) {};
			\node (R6) at (-1.45,1.6) {};
			\node (R7) at (-1.95,.3) {};
			\node (R8) at (0.4,-.9) {};
			\node[inner sep =1.] (RS) at (.55,.725) {};
			\draw[fill=black!10!white, thick] (R5.center) -- (R6.center) -- (R7.center) -- (R8.center) --cycle;
			\draw[fill=black!20!white, thick] (R1.center) -- (R5.center) -- (R8.center) -- (R4.center) --cycle;
			\draw[fill=black!30!white, thick] (R7.center) -- (R8.center) -- (R4.center) -- (R3.center) --cycle;
			\draw[dashed] (R1.center) -- (R2.center) -- (R3.center) -- (R4.center) --cycle;
			\draw[dashed] (R2.center) -- (R3.center) -- (R7.center) -- (R6.center) --cycle;
		\end{scope}

		\draw[semithick, dashed]  (T0) -- (T3);
		\draw[semithick, dashed]  (T0) -- (T2);
		
		\begin{scope}[scale=1.5]
			\node (T1) at (0.75,-1.2) {};
			\node (T2) at (-0.75,1.2) {};
			\node (T3) at (1.2,.9) {};
			\node (T4) at (-1.2,-.9) {};
			\node (T5) at  (1.95,-.3) {};
			\node (T6) at (.45,2.1) {};
			\node (T7) at  (-1.95,.3)  {};
			\node (T8) at (-.45,-2.1) {};
			
			\draw[dashed, ] (T5.center) -- (T6.center) -- (T7.center) -- (T8.center) --cycle;
			\draw[dashed, very thin]  (T1.center) edge (T0.center);
			\draw[dashed, very thin]  (T0.center) edge (T4.center);
		\end{scope}
		
		\draw[fill=black]  (T0) ellipse (0.05 and 0.05);
			
		\begin{scope}[scale=1.2]
			\draw [-latex, very thin] (0,0)--(.5,1.3) {};
			\draw [-latex, very thin] (0,0)--(1.2,-.6) {};
			\draw [-latex, very thin] (0,0)  {}--(-.8,.56) {};
			\node at (0.75,1.2) {$y_R$};
			\node at (1.35,-.45) {$x_R$};
			\node at (-.75,0.3) {$z_R$};
		\end{scope}		
		
		\node at (1.4,.6) {$\idbf{}{v}{m}$};
		\node at (-.5,-1.25) {$\id{}{\omega}{m}$};
		\draw[-latex, thick] (T0) -- (1.75,.4);
		\draw[-latex, thick] plot[smooth, tension=.9] coordinates {(-1.4,-.4) (-.8,-1.) (.,-1.)};

		\draw [dashed, very thin] (0,0)--(2,-1.) {};
		\draw[-latex]  plot[smooth, tension=.9] coordinates {(1.7,1.25) (2,.1) (1.95,-.98)};
		\node at (2.3,-0.2) {$\id{}{\gamma}{R}$};
	\end{scope}
	
	\node (W0) at (-2.5,0.5) {};
	\begin{scope}[scale=1.5, rotate around={0:(W0)}, shift={(W0)}]
		\draw [-latex, very thin] (0,0)--(1,0);
		\draw [-latex, very thin] (0,0)--(0,1);
		\draw [-latex, very thin] (0,0) node (v2) {}--(0.4,-0.6);
		\node at (-0.3,0) {$(W)$};
		\node at (.2,-.6) {$x$};
		\node at (1,-.2) {$y$};
		\node at (-0.2,1) {$z$};
	\end{scope}
	
	\node (M) at (0,0) {};
	\begin{scope}[scale=.7, rotate around={0:(M)}, shift={(M)}]

		\node (M1) at (0,0) {};
		\node (M2) at (-4,6) {};
		\node (M3) at (1.5,0) {};
		\node (M4) at (-2.5,6) {};
		\node (M5) at (4.5,2.5) {};
		\node (M6) at (0.5,8.5) {};
		\node (M7) at (8.5,2) {};
		\node (M8) at (4.5,8) {};
		
		\draw[dashed]  plot[smooth, tension=.7, dotted] coordinates {(0,0) (1.5,0) (4.5,2.5) (8.5,2) };
		\begin{scope}[scale=1, shift={(-2,3)}]
			\draw[very thin,dotted]  plot[smooth, tension=.7] coordinates {(0,0) (1.5,0) (4.5,2.5) (8.5,2) };
		\end{scope}
		\begin{scope}[scale=1, shift={(-4,6)}]
			\draw [dashed]  plot[smooth, tension=.7, dotted] coordinates {(0,0) (1.5,0) (4.5,2.5) (8.5,2) };
		\end{scope}
		\draw (M2.center) edge (M1.center);
		\draw[very thin, dotted]  (M4.center) edge (M3.center);
		\draw[very thin, dotted]  (M6.center) edge (M5.center);
		\draw[very thin, dotted]  (M8.center) edge (M7.center);
		\node[] at (1.2,1.8) {$\TpM$};
	\end{scope}
	
	\draw[fill=col_orange!35!white]  plot[smooth cycle, tension=.7] coordinates {(1.5,-1) (4,-1)  (5.,-0.5) (5,0.5) (3.75,1.25)  (2.5,1) (2.7,.5)} ;
	\node (U0) at (4.5,0) {};
	\begin{scope}[scale=1, rotate around={0:(U0)}, shift={(U0)}]
		
		\begin{scope}[scale=2, shift={((0,0)}]
			\draw [-latex,very thin] (0,0)--(-.75,.5);
			\draw [-latex,very thin] (0,0)--(-0.5,-0.45);
			\node at (0.15,0) {$\U$};
			\node at (-.6,.5) {$v$};
			\node at (-.3,-.4) {$u$};
		\end{scope}
		\node (t) at (-1.5,.1) {};
		\node at (-1.35,-.25) {$\idbf{}{t}{R}$};
		
		\draw[fill=black]  (t) ellipse (0.05 and 0.05);
	\end{scope}
	
	\node[fill=black!10!white, inner sep = .2] at (.96,3.95) {$\idbf{}{p}{R}$};
	\node at (5,2) {$\M$};
	\node at (3.75,5.75) {$\idbf{}{B}{1}$};
	\node at (-0.15,6.25) {$\idbf{}{B}{2}$};
	\draw[-latex,  thick, color=orange]  plot[smooth, tension=.7] coordinates {(1.5,3.5) (2,3) (3.05,0.4)};
	\draw[latex-,  thick, color=orange]  plot[smooth, tension=.7] coordinates {(1.1,3.1) (2.1,0.6) (2.65,0.)};
	\node at (1.8,0.3) {\textcolor{orange}{$\sigma^{-1}$}};
	\node at (2.9,1.4) {\textcolor{orange}{$\sigma$}};
\end{tikzpicture}
	\caption{A manifold $\M$, the chart $\U$ and the tangent space $\TpM$ at the robot position $\idbf{}{p}{R} = \sigma^{-1}(\idbf{}{t}{R}) \in \M$, illustrating the proposed approach of estimating $\idbf{}{t}{R}$ on the chart and the heading $\id{}{\gamma}{R}$ within the tangent space, while the odometry velocities $\idbf{}{v}{m}{},\id{}{\omega}{m}{}$ are measured in the tangent plane.}
	\label{fig:manifold}
\end{figure}
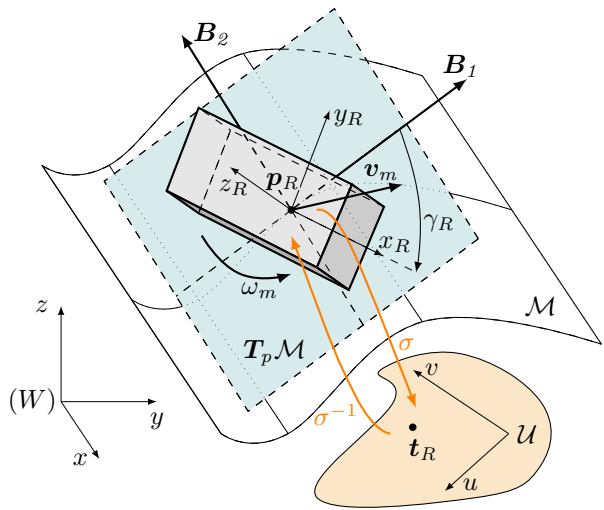

\subsection{Smooth Surfaces}
Providing suitable terrain approximations for diverse applications of UGVs, general smooth surfaces embedded in $\RIII$ can be defined via implicit equations $\Phi\left(x,y,z\right) = 0$, with functions $\Phi: \RIII \rightarrow \RI$ describing their geometry. As noted above, they formally represent parallelizable 2-dimensional  manifolds $\M = \{(x,y,z) \in \RIII\;|\;\Phi\left(x,y,z\right) = 0\}$ illustrated in Fig. \ref{fig:manifold}. Consequently, a non-unique basis of $\TpM$ at each $\bb{p} \in \M$ can generally be found analytically. For the discussed application, an orthonormal parallelization is preferred, as two such tangent vectors $\idbf{}{B}{1}(\bb{p}), \idbf{}{B}{2}(\bb{p})$ together with the surface normal $\idbf{}{N}{}(\bb{p})$ in $\RIII$ form the rotation matrix
\begin{gather}
	\idbf{}{R}{W\TpM} = \left[ \idbf{}{B}{1}(\bb{p}) \;|\; \idbf{}{B}{2}(\bb{p}) \;|\; \idbf{}{N}{}(\bb{p}) \right],
\end{gather}
expressing the orientation of the tangent space within the world frame $(W)$. While such a representation can be given for all parallelizable 2-manifolds, a limitation to explicit surfaces allows for a convenient formulation via the directional derivatives of $\Phi$. These surfaces are a subgroup of implicit geometries, where one variable, for instance the elevation $z$, can be written as a function of the remaining two
\begin{gather}
	\Phi\left(x,y,z\right) = S(x,y) - z = 0\rightarrow z = S(x,y).
\end{gather}
In this specific case, elementary rotations around the x- and y-axis derived from the directional derivatives $D_x, D_y$
\begin{gather}
	\idbf{}{R}{x} = \frac{1}{\sqrt{1+(D_y\Phi)^2}}\begin{bmatrix} 
		1 & 0 & 0 \\
		0 & 1 & -D_y\Phi \\
		0 & D_y\Phi & 1 \\	
	\end{bmatrix}\hspace{.9ex} \\
	\idbf{}{R}{y} = \frac{1}{\sqrt{1+(D_x\Phi)^2}}\begin{bmatrix} 
		1 & 0 & -D_x\Phi \\
		0 & 1 & 0 \\
		D_x\Phi & 0 & 1 \\	
	\end{bmatrix},
\end{gather}
 can be leveraged to compose $\idbf{}{R}{W\TpM}$ based on Tait–Bryan rotations. Its columns are thus forming a new orthonormal basis of $\TpM$
 \begin{gather}
 	\idbf{}{R}{W\TpM} = \idbf{}{R}{x}\idbf{}{R}{y} = \left[ \idbf{}{B}{1}'(\bb{p}) \;|\; \idbf{}{B}{2}'(\bb{p}) \;|\; \idbf{}{N}{}'(\bb{p})\right].
 \end{gather} 
 This particular parallelization simplifies the necessary formulations, as now the robot's global orientation $\idbf{}{R}{WR}$ can be separated into components introduced by the manifold $\idbf{}{R}{W\TpM}$ and the heading $\idbf{}{R}{\TpM R}(\id{}{\gamma}{R})$ within the tangent plane, an elementary rotation along the local z-axis by $\id{}{\gamma}{R}$,
 \begin{gather}
 	\idbf{}{R}{WR} = \idbf{}{R}{W\TpM}\idbf{}{R}{\TpM R}.
 \end{gather}

\subsection{State Definition and Dynamics}
Given these results, the vehicle's pose is fully parameterized by its location on the chart $\idbf{}{t}{R}$ and its heading angle $\id{}{\gamma}{R}$ w.r.t. the tangent space, simplifying the estimation task to the product space $\U\times\SOII$. For modeling the corresponding dynamics, the linear and angular odometry velocities $\idbf{R}{v}{m} \in \RII$, $\id{}{\omega}{m}\in \RI$ measured in the robot frame $(R)$, are considered as system inputs $\idbf{}{u}{}$. Given as locally 2-dimensional measurements within $\TpM$, the linear components are mapped onto the chart, to accurately model the motion along the manifold. Assuming additive input measurement noise $\idbf{}{n}{v} \sim \mathcal{N}(\bb{0}, \idbf{}{\sigma}{v})$ and $\id{}{n}{\omega} \sim \mathcal{N}(0, \id{}{\sigma}{\omega})$, the state vector $\idbf{}{x}{}$ and its dynamics $\idbfdot{}{x}{}$ can be written as
\begin{gather}
	\idbf{}{x}{} = \begin{bmatrix} \idbf{}{t}{R}^T &\id{}{\gamma}{R}\end{bmatrix}^T, \quad
	\idbf{}{u}{} = \begin{bmatrix} \idbf{}{v}{m}^T &\id{}{\omega}{m}\end{bmatrix}^T 
	\\
	\idbfdot{}{t}{R} = \left.\pfpx{\sigma}{\idbf{}{p}{}} \right|_{\func{\sigma^{-1}}{\idbf{}{t}{R}}} \idbf{}{R}{W\TpM} \idbf{}{R}{\TpM R} \begin{bmatrix} \idbf{R}{v}{m} - \idbf{}{n}{v}\\ 0 \end{bmatrix} \label{eq:lin_chart_vel}
	\\
	\iddot{}{\gamma}{R} = \id{}{\omega}{m} - \id{}{n}{\omega}.
\end{gather}
Here, the sign of the noise terms may be chosen arbitrarily and is selected for convenience. It is important to note that \eqref{eq:lin_chart_vel} applies a linear mapping to the measured velocities, preserving the gaussianity of their input noise.

Formulating an ESEKF, as introduced by \cite{roumeliotis1999}, for this model, the error states $\delta\idbf{}{x}{}$ are defined via the difference between the true, perturbed states $\idbf{}{x}{}$ and the nominal, noise-free ones $\idbf{}{x}{n}$. Hence, the error dynamics $\delta\idbfdot{}{x}{}$ necessary for the filter formulation are given by
\begin{gather}
	\idbf{}{x}{n} = \left.\idbf{}{x}{}\right|_{\idbf{}{n}{v}=\bb{0},\id{}{n}{\omega}=0}
	\\
	\delta\idbf{}{x}{} = \idbf{}{x}{} - \idbf{}{x}{n} = \begin{bmatrix} \delta\idbf{}{t}{R}^T & \delta\id{}{\theta}{R}\end{bmatrix}^T
	\\
	\delta\idbfdot{}{x}{} = \begin{bmatrix}\delta\idbfdot{}{t}{R}^T & \delta\iddot{}{\theta}{R}\end{bmatrix}^T,
\end{gather}
where $\delta\idbf{}{t}{R}$ denotes the positional error and $\delta\id{}{\theta}{R}$ the rotational one. These can be linearized regarding $\delta\idbf{}{x}{}$ for a given state $\idbf{}{x}{n}$  by applying a first-order Taylor approximation to attain a system linear w.r.t. the error states and the input noise, as required for implementing the indirect filtering method. 

\begin{figure}[b]
	\centering
	\begin{tikzpicture}[scale=1.2]
	
	\node[inner sep=.1] (W0) at (.7,1.8) {};
	\begin{scope}[scale=1, rotate around={0:(W0)}, shift={(W0)}]
		\draw [-latex, very thin] (0,0)--(.5,0);
		\draw [-latex, very thin] (0,0)--(0,.5);
		\node at (.2,-.3) {$(W)$};
	\end{scope}

	\node[inner sep=.1] (U0) at (1,1) {};
	\begin{scope}[scale=1, rotate around={0:(U0)}, shift={(U0)}]
		\draw [-latex, very thin] (0,0)--(1,-.5);
		\draw [-latex, very thin] (0,0)--(.9,.3);
		\node at (-0.,-.3) {$\U$};
		\node[inner sep=1.5] (t0) at (1,0) {};
		\node[inner sep=1.5]  (tm) at (1.83,-.085) {};
		\node[inner sep=1.5]  (t1) at (3,-.2) {};
		
		\draw[fill=black]  (t0) ellipse (0.05 and 0.05);
		\draw[fill=black]  (t1) ellipse (0.05 and 0.05);
		
		\node[] at (1,-.2) {$\idbf{}{t}{R,k}$};
		\node[] at (3.2,-.4) {$\idbf{}{t}{R,k} + \Delta \idbf{}{t}{R,k}$};
	\end{scope}
	
	\node[inner sep=.1] (R0) at (2,2.53) {}; 
	\begin{scope}[scale=1, rotate around={35:(R0)}, shift={(R0)}]
		
		\draw[thick, fill=black!10!white]  (-.5,0) rectangle (.5,.5);
		\draw[fill=black]  (R0) ellipse (0.05 and 0.05);
		\node[inner sep=.1] (S0) at (.5,1) {};
		\draw[fill=black]  (S0) ellipse (0.05 and 0.05);
		\draw[thick] (.5,.5) -- (S0);
		
		\draw[thick, -latex, color = teal!80!white] (R0) -- (1,0);
		\draw[-latex, color = teal!80!white]  plot[smooth, tension=.9] coordinates {(0.3,0.875) (0,.75) (0.5,0.6)  (1,.7) (0.65,0.85)};
		
		\node at (0,.25) {$\idbf{}{x}{k}$};
		\node[color = teal!80!white] at (1,.2) {$\idbf{}{v}{m,k}$};
		\node[color = teal!80!white] at (-.35,.9) {$\id{}{\omega}{m,k}$};
		\node[] (vm) at (1,0) {};
	\end{scope}
	
	\node[inner sep=.1] (R2) at (4,2.85) {}; 
	\begin{scope}[scale=1, rotate around={-13:(R2)}, shift={(R2)}]
		
		\draw[fill=black!15!white]  (.45,0) rectangle (.55,.5);
		\draw[fill=black!5!white]  (-.55,0) rectangle (.45,.5);
		\draw[dashed]  (-.45,0) rectangle (-.55,.5);
		\draw[fill=black, color = black]  (R2) ellipse (0.05 and 0.05);
		\node[inner sep=.1] (S0) at (.45,1) {};
		\draw[fill=black]  (S0) ellipse (0.05 and 0.05);
		\draw (.45,.5) -- (S0);
		
		\node at (0,.25) {$\idbf{}{x}{k+1}$};
	\end{scope}
	
	\node[inner sep=.1] (R3) at (5.8,2.5) {}; 
	\begin{scope}[scale=1, rotate around={5:(R3)}, shift={(R3)}]
		
		\draw[dashed, color=col_sage!60!black]  (.45,0) rectangle (.55,.5);
		\draw[dashed, color=col_sage!60!black]  (-.55,0) rectangle (.45,.5);
		\node[dashed, color=col_sage!60!black] (S0) at (.45,1) {};
		\draw[color=col_sage!60!black, fill =col_sage!60!black]  (S0) ellipse (0.05 and 0.05);
		\draw[dashed, color=col_sage!60!black] (.45,.5) -- (S0);
		\draw[color=col_sage!60!black, fill=col_sage!60!black]  (R3) ellipse (0.05 and 0.05);
		\draw[, -latex] (R3.center) -- (S0.center);
	\end{scope}
	
	\node[inner sep=.1] (A0) at (2,4.5) {};
	\begin{scope}[scale=1, rotate around={45:(A0)}, shift={(A0)}]
		\draw[fill=col_teal!50!black, color= col_teal!50!black]  (0,0) node (v1) {}  ellipse (0.05 and 0.05);
		\draw[draw=col_teal!50!black] (-.125,-.125) rectangle ++(0.25,0.25);
		\node[rotate=0, color = col_teal!50!black] at (-.4,0) {$A$};
		\begin{scope}[rotate around={-40:(A0)}]
			\draw[dotted, color=col_sage!60!black, ] (4.27,0) arc (0:-50:4.27);
		\end{scope}
	\end{scope}
	
	\draw[]  plot[smooth, tension=.7] coordinates {(1,2) (1.4,2.1) (3,2.95) (5.5,2.5) (6.55,2.75)};
	
	\node at (4.7,2.4) {$\M$};
	\draw[dashdotted, very thin]  (R0.center) -- (t0);
	\draw[dotted,]  (vm.center) -- (tm.center);
	\draw[dashdotted, very thin]  (R2) -- (t1.center);
	\node at (1.8,1.7) {$\sigma$};
	\node at (4.35,1.7) {$\sigma^{-1}$};
	
	\node[color=orange] at (2.45,1.25) {$\idbfdot{}{t}{R_k}$};
	\draw[dotted,] (t0) -- (t1);
	\draw[thick, -latex,color=orange]  (t0.center) -- (tm.center);
	\node[color=col_sage!60!black] at (5.65,3.5) {$\idbf{}{z}{p}, \idbf{}{z}{q}$};
	\node (v2) at (6.25,4.5) {};
	\draw[latex-latex, dashed, color=col_sage!60!black, ]  (v1.center) -- (v2.center);
	\node[color=col_sage!60!black, ] at (4,4.25) {$\id{}{z}{d}$};
	\node at (1.6,3.7) {$S$};
	\node at (5.57,2.75) {$\idbf{}{r}{RS}$};
\end{tikzpicture}
	\caption{Depiction of the state propagation model and the proposed velocity mapping, together with the considered position, orientation and distance measurements, $\idbf{}{z}{p}$, $\idbf{}{z}{q}$ and $\id{}{z}{d}$ respectively. $S$ denotes the on-board sensor with a local offset $\idbf{}{r}{RS}$ while $A$ indicates the range measurement anchor.}
	\label{fig:measurements}
\end{figure}
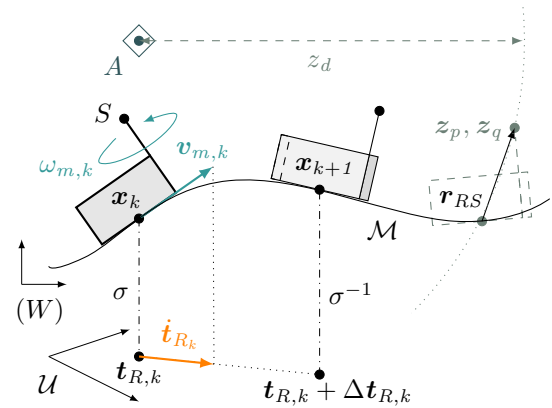

\subsection{State Propagation}
While the ESEKF tracks $\delta\idbf{}{x}{}$, the vehicle's pose is propagated based on the measured velocities, before applying the filter corrections. Unlike prior work that formulates this step in world coordinates \cite{starbuck2021}, we propose a computation directly in the reduced state space. Via the same linear mapping of odometry data as outlined in \eqref{eq:lin_chart_vel}, the state displacement model for discrete-time propagation is hence expressed as
\begin{gather}
	\Delta \idbf{}{t}{R,k} = \left.\pfpx{\sigma}{\idbf{}{p}{}} \right|_{\func{\sigma^{-1}}{\idbf{}{t}{R,k}}} \idbf{}{R}{WR,k} \begin{bmatrix} \idbf{R}{v}{m,k}\\ 0 \end{bmatrix} \Delta t \label{eq:lin_chart_dis}
	\\
	\Delta  \idbf{}{\gamma}{R,k} = \id{}{\omega}{m,k}  \Delta t,
\end{gather}
at a given time step $k$ and with a sample time $\Delta t$. 

\subsection{Measurement Models in $\RIII$}
To correct the filter’s estimates, two types of measurements providing global localization references are studied for this application. The first involves global pose measurements consisting of the measured position $\idbf{}{z}{p}$ and quaternion orientation $\idbf{}{z}{q}$ of an on-board sensor $S$ w.r.t. the world frame. Such data can be generated for instance by scan-matching-based LiDAR localization systems \cite{chalvatzaras2023}. Accounting for the on-board device offset $\idbf{R}{r}{RS}$, $\idbf{}{q}{RS}$, these loosely-coupled inputs are defined via
\begin{gather}
	\idbf{}{z}{p} = \func{\sigma^{-1}}{\idbf{}{t}{R}} +\idbf{}{R}{W\TpM} \idbf{}{R}{\TpM R} \idbf{R}{r}{RS} 
	\\
	\idbf{}{z}{q} = \idbf{}{q}{}\left\{\idbf{}{R}{W\TpM} \right\} \otimes \idbf{}{q}{}\left\{\idbf{}{R}{\TpM R}\right\}\otimes\idbf{}{q}{RS},
\end{gather}
where the operator $\bb{q}\left\{\bb{R}\right\}$ denotes the quaternion corresponding to a rotation matrix $\bb{R}$ and $\otimes$ represents quaternion multiplication. As $\idbf{}{R}{W\TpM}$ depends on $\idbf{}{t}{R}$, the orientation data $\idbf{}{z}{q}$ also encodes information on the vehicle's location.

The second model incorporates ranging data, such as that produced by UWB systems. These measurements represent the Euclidean distance between an on-board sensor $S$ and a static anchor $A$ located at $\idbf{W}{r}{A}$, as shown in Fig. \ref{fig:measurements} 
\begin{gather}
	\id{}{z}{} = \left\|\func{\sigma^{-1}}{\idbf{}{t}{R}} + \idbf{}{R}{W\TpM} \idbf{}{R}{\TpM R} \idbf{R}{r}{RS}  - \idbf{W}{r}{A}\right\|_2.
\end{gather}

Like the error state dynamics, these measurement models can be linearized w.r.t. $\delta\idbf{}{x}{}$ via first-order Taylor expansions and the small-angle approximations detailed in \cite{sola2017}.

\subsection{Projected Measurement Models on $\U$}
The measurement models described above, though analytically valid, may misrepresent the actual uncertainties by neglecting the surface constraints for covariance modeling. To address this, we develop strategies for mapping the position and distance data, $\idbf{}{z}{p}$ and $\id{}{z}{d}$, onto the chart while preserving the Gaussian nature of their noise distributions.

The robot position corresponding to a generic pose measurement of an on-board sensor $\idbf{W}{r}{S,m}$ does generally not lie on the manifold and therefore cannot be mapped to $\U$ directly. However, by accounting for the device offset $\idbf{R}{r}{RS}$ and the current orientation estimate $\idbfhat{}{R}{WR}$, the sensor data can be associated with $\M$ by finding the closest point on the surface, $\idbf{}{z}{p,\M} \in \M$. This result is then mapped to the chart via $\sigma$ to formulate the new 2-dimensional position measurement $\idbf{}{z}{t}\in\U$. While specific terrain models may allow for analytic solutions, this process generally requires numerical optimization
\begin{gather}
	\idbf{}{z}{t} = \sigma\left(\arg \min_{\bb{p}\in\M}\left\|\idbf{W}{r}{S,m} + \idbfhat{}{R}{WR}\idbf{R}{r}{RS} - \bb{p}\right\|_2\right). \label{eq:zt}
\end{gather}

\begin{figure}[h]
	\centering
	\begin{tikzpicture}[scale=.65]
	
	\clip  (-4.2,-1.8) rectangle (6.7,6.8);
	
	\node (M) at (0,0) {};
	\begin{scope}[scale=.7, rotate around={0:(M)}, shift={(M)}]
		
		\node (M1) at (0,0) {};
		\node (M2) at (-4,6) {};
		\node (M3) at (1.5,0) {};
		\node (M4) at (-2.5,6) {};
		\node (M5) at (4.5,2.5) {};
		\node (M6) at (0.5,8.5) {};
		\node (M7) at (8.5,2) {};
		\node (M8) at (4.5,8) {};
		
		\draw  plot[smooth, tension=.7] coordinates {(0,0) (1.5,0) (4.5,2.5) (8.5,2)};
		\begin{scope}[scale=1, shift={(-2,3)}]
			\draw[very thin]  plot[smooth, tension=.7] coordinates {(0,0) (1.5,0) (4.5,2.5) (8.5,2)};
		\end{scope}
		\begin{scope}[scale=1, shift={(-4,6)}]
			\draw  plot[smooth, tension=.7] coordinates {(0,0) (1.5,0) (4.5,2.5) (8.5,2)};
		\end{scope}
		\draw (M2.center) edge (M1.center);
		\draw[very thin]  (M4.center) edge (M3.center);
		\draw[very thin]  (M6.center) edge (M5.center);
		\draw[]  (M8.center) edge (M7.center);
	\end{scope}

	\node[inner sep=.1] (T0) at (1.1,3.5) {}; 
	\begin{scope}[scale=1, rotate around={0:(T0)}, shift={(T0)}]
		
		\begin{scope}[scale=1.5]

			\node (T1) at (0.75,-1.2) {};
			\node (T2) at (-0.75,1.2) {};
			\node (T3) at (1.2,.9) {};
			\node (T4) at (-1.2,-.9) {};
			\node (T5) at  (1.95,-.3) {};
			\node (T6) at (.45,2.1) {};
			\node (T7) at  (-1.95,.3)  {};
			\node (T8) at (-.45,-2.1) {};
			
			\draw[dashed, fill=col_teal!25!white] (T5.center) -- (T6.center) -- (T7.center) -- (T8.center) --cycle;
		\end{scope}
		
		\node (S0) at (0,0) {};
		\begin{scope}[scale=1, rotate around={0:(S0)}, shift={(S0)}]	

			\node (e1) at  (1.17,-.15) {};
			\node (e2) at (-1.17,.15) {};	
			\node (e3) at  (.3,1.4) {};
			\node (e4) at (-.3,-1.4) {};
			
			\node (e11) at  (1.145,-.3) {};
			\node (e22) at (-1.145,.3) {};	
			\node (e33) at  (.35,1.39) {};
			\node (e44) at (-.35,-1.39) {};

			\draw[dashed,color=orange] (e1.center) -- (e2.center);
			\draw[dashed, color=orange]  (e3.center) -- (e4.center);
			\draw[ color = orange, thick]  plot[smooth cycle, tension=1] coordinates {(e11) (e33) (e22) (e44)};
			\draw[-latex,orange, thick] (S0.center) -- (e1.center);
			\draw[-latex,orange, thick] (S0.center) -- (e3.center);
			
			\node (s1) at  (1.2,0) {};
			\node (s2) at (-1.2,0) {};	
			\node (s3) at  (0,2) {};
			\node (s4) at (0,-2) {};
			\node (s5) at  (-.1,.3) {};
			\node (s6) at (.1,-.3) {};
			
			\node (s11) at  (1.2,0) {};
			\node (s22) at (-1.2,0) {};	
			\node (s33) at  (0,2) {};
			\node (s44) at (0,-2) {};
			
			\node (s55) at  (.11,1.725) {};
			\node (s66) at (-.11,-1.725) {};
			\node (s77) at  (-.2,1) {};
			\node (s88) at (.2,-1) {};
			
			\node (s99) at (1.19,.04) {};
			\node (s101) at (-1.19,-.04) {};
			\node (s111) at (-.2,.3) {};
			\node (s121) at (.2,-.3) {};
			
			\draw[thick] (.235,0) arc(-0:90:.235cm and 2cm);
			\draw[dashed] (.235,0) arc(0:137:.235cm and 2cm);
			\draw[thick] (.235,0) arc(-0:-44.5:.235cm and 2cm);
			\draw[dashed, very thin] (-.235,0) arc(0:43:-.235cm and 2cm);
			\draw[dashed, very thin] (-.235,0) arc(0:-135.5:-.235cm and 2cm);
			
			\draw[thick] (1.2,0) arc(-0:180:1.2cm and 2cm);
			\draw[thick] (1.2,0) arc(-0:-4.5:1.2cm and 2cm);
			\draw[dashed, very thin] (1.2,0) arc(0:-195:1.2cm and 2cm);
			
			\draw[dashed, very thin] (1.2,0) arc(-0:182:1.2cm and .5cm);
			\draw[thick] (1.2,0) arc(0:-197:1.2cm and .5cm);

			\draw[fill=black]  (0,0) node (v1) {} ellipse (0.075 and 0.075);
			
		\end{scope}
	\end{scope}

	\node (W0) at (-3,.5) {};
	\begin{scope}[scale=1.5, rotate around={0:(W0)}, shift={(W0)}]
		\draw [-latex, very thin] (0,0)--(1,0);
		\draw [-latex, very thin] (0,0)--(0,1);
		\draw [-latex, very thin] (0,0) node (v2) {}--(0.4,-0.6);
		\node at (-0.35,0) {$(W)$};
		\node at (.2,-.6) {$x$};
		\node at (1,-.2) {$y$};
		\node at (-0.2,1) {$z$};
	\end{scope}

	\node (M) at (0,0) {};
	\begin{scope}[scale=.7, rotate around={0:(M)}, shift={(M)}]

		\node (M1) at (0,0) {};
		\node (M2) at (-4,6) {};
		\node (M3) at (1.5,0) {};
		\node (M4) at (-2.5,6) {};
		\node (M5) at (4.5,2.5) {};
		\node (M6) at (0.5,8.5) {};
		\node (M7) at (8.5,2) {};
		\node (M8) at (4.5,8) {};

		\draw[dashed]  plot[smooth, tension=.7, dotted] coordinates {(0,0) (1.5,0) (4.5,2.5) (8.5,2)};
		\begin{scope}[scale=1, shift={(-2,3)}]
			\draw[very thin,dotted]  plot[smooth, tension=.7] coordinates {(0,0) (1.5,0) (4.5,2.5) (8.5,2)};
		\end{scope}
		\begin{scope}[scale=1, shift={(-4,6)}]
			\draw [dashed]  plot[smooth, tension=.7, dotted] coordinates {(0,0) (1.5,0) (4.5,2.5) (8.5,2)};
		\end{scope}
		\draw (M2.center) edge (M1.center);
		\draw[very thin, dotted]  (M4.center) edge (M3.center);
		\draw[very thin, dotted]  (M6.center) edge (M5.center);
		\draw[very thin, dotted]  (M8.center) edge (M7.center);
	\end{scope}

	\node (U0) at (0,-.25) {};
	\begin{scope}[scale=1, rotate around={0:(U0)}, shift={(U0)}]
		
		\begin{scope}[scale=2, shift={((3,-.)}]
			\draw [-latex,very thin] (0,0)--(-.9,.3);
			\draw [-latex,very thin] (0,0)--(-.5,-0.7);
			\node at (.2,0) {$\U$};
			\node at (-.7,0.4) {$v$};
			\node at (-.2,-.6) {$u$};
		\end{scope}
		
		\node (t0) at (3.1,-.5) {};
		\begin{scope}[scale=1, rotate around={-17:(t0)}, shift={(t0)}]
			
			\node (t1) at  (1.17,.1) {};
			\node (t2) at (-1.17,-.1) {};	
			\node (t3) at  (.3,.8) {};
			\node (t4) at (-.3,-.8) {};
			
			\node (t11) at  (1.13,-.2) {};
			\node (t22) at (-1.13,.2) {};	
			\node (t33) at  (.35,.8) {};
			\node (t44) at (-.35,-.8) {};
			
			\draw[dashed,color=orange] (t1.center) -- (t2.center);
			\draw[dashed, color=orange]  (t3.center) -- (t4.center);
			\draw[ color = orange, thick]  plot[smooth cycle, tension=1] coordinates {(t11) (t33) (t22) (t44)};
			\draw[color=orange,dashdotdotted, very thin]  (e1.center) edge (t1.center);	
			\draw[color=orange,dashdotdotted, very thin]  (e2.center) edge (t2.center);	
			\draw[color=orange,dashdotdotted, very thin]  (e3.center) edge (t3.center);
			\draw[color=orange,dashdotdotted, very thin]  (e4.center) edge (t4.center);

			\draw[-latex,orange, thick] (t0.center) -- (t1.center);
			\draw[-latex,orange, thick] (t0.center) -- (t3.center);
			\draw[fill=black]  (0,0) ellipse (0.075 and 0.075);
		\end{scope}
		
		\node at (4.5,-.1) {\textcolor{orange}{$\idbf{}{P}{t}$}};
		\node at (3.3,-.85) {$\idbf{}{z}{t}$};
		
	\end{scope}
	
	\node[inner sep=.1] (R0) at (-2.,5) {}; 
	\begin{scope}[scale=1, rotate around={45:(R0)}, shift={(R0)}]
		
		\draw[dashed, fill=black!10!white]  (-.5,0) rectangle (.5,.5);
		\draw[]  (R0) ellipse (0.075 and 0.075);
		\node[inner sep=.1] (RS0) at (.5,1) {};
		\draw[fill=black]  (RS0) ellipse (0.075 and 0.075);
		\draw[dashed] (.5,.5) -- (RS0);
		
		\draw[-latex]  (R0) edge (RS0);
		
		\node at (1.4,.2) {$\idbf{}{r}{S,m}, \idbf{}{P}{m}$};
		\node at (-.2,1) {$\idbf{}{r}{RS}$};
	\end{scope}
	
	\node at (6.25,1.8) {$\M$};
	\node at (2,5.5) {$\idbf{}{P}{\M}$};
	\node[color = orange] at (.85,4.3) {$\idbf{}{r}{1}$};
	\node[color = orange] at (2.7,3.4) {$\idbf{}{r}{2}$};
	\node at (-.8,3.5) {$\idbf{}{z}{p,\M}$};
	\draw[-latex]  plot[smooth, tension=.9] coordinates {(-0.5,3.3) (0.25,3) (1.,3.4)};
	\draw[dashdotted] (R0.center) -- (S0.center) ;
	\draw[dashdotted] (S0.center) -- (t0.center) ;
	\node[] at (.7,1.15) {$\TpM$};
\end{tikzpicture}
	\caption{The proposed method for projecting position measurements $\idbf{}{r}{S,m}$ and their covariances $\idbf{}{P}{m}$ onto the chart by intersecting the covariance ellipsoid $\idbf{}{P}{\M}$ of the associated point $\idbf{}{z}{p,\M}\in\M$ with the tangent plane and mapping the conjugate diameter vectors of the resulting ellipse.}
	\label{fig:proj_position}
\end{figure}
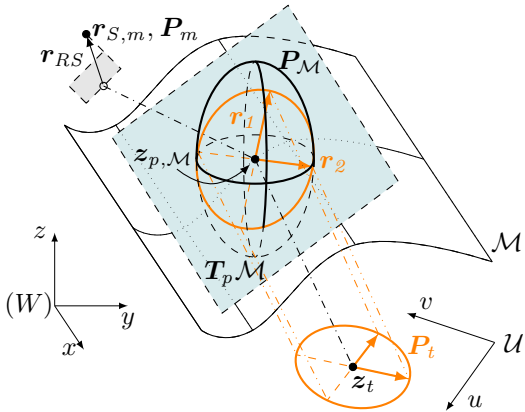

To approximate the true uncertainty along the surface, the corresponding covariance ellipsoid $\idbf{}{P}{\M}$ is intersected with the local tangent plane at $\idbf{}{z}{p,\M}$. Since \eqref{eq:zt} incorporates data from the current estimate, the measurement covariance matrix $\idbf{}{P}{m}$ of the sensor is augmented with the current state uncertainty $\idbf{}{P}{x}$ via $\idbf{}{J}{m\M}$, the Jacobian of the shift w.r.t. $\bb{x}$. The resulting intersection ellipse can then be parameterized by its conjugate diameter vectors $\idbf{W}{r}{1}, \idbf{W}{r}{2}$ obtained through eigenvalue decomposition. Analogous to the mapping of linear velocities, see \eqref{eq:lin_chart_vel}, these are mapped onto the chart to model the 2-dimensional uncertainty $\idbf{}{P}{t}$ of $\idbf{}{z}{t}$ on $\U$, as illustrated in Fig. \ref{fig:proj_position}
\begin{gather}
	\idbf{}{P}{\M} = \idbf{}{P}{m}  + \idbf{}{J}{m\M} \idbf{}{P}{x} \idbf{}{J}{m\M}^T
	\\
	\idbf{}{P}{t}^{-1} = \left(\pfpx{\sigma}{\bb{p}}\left[\idbf{W}{r}{1} \;|\; \idbf{W}{r}{2}\right]\right)\left(\pfpx{\sigma}{\bb{p}}\left[\idbf{W}{r}{1} \;|\; \idbf{W}{r}{2}\right]\right)^T.
\end{gather}

\begin{figure}[t]
	\centering
	\vspace{1ex}
	\begin{tikzpicture}[scale=.7]
	\clip  (-1.9,-.2) rectangle (8.1,4.25);
	\node[inner sep=.1] (W0) at (-1,1.3) {};
	\begin{scope}[scale=1, rotate around={0:(W0)}, shift={(W0)}]
		\draw [-latex, very thin] (0,0)--(1,0);
		\draw [-latex, very thin] (0,0)--(0,1);
		\node at (-0.45,.2) {$(W)$};
	\end{scope}
	
	\node[inner sep=.1] (U0) at (-.5,.5) {};
	\begin{scope}[scale=1, rotate around={0:(U0)}, shift={(U0)}]
		\draw [-latex, very thin] (0,0)--(1,-.5);
		\draw [-latex, very thin] (0,0)--(.9,.3);
		\node at (-0.2,.2) {$\U$};
		\node[inner sep=1.5] (t0) at (2.5,.04) {};	
	\end{scope}
	
	\node[inner sep=.1] (R0) at (2,2.53) {}; 
	\begin{scope}[scale=1, rotate around={36:(R0)}, shift={(R0)}]
		
		\draw[fill=black!10!white]  (-.5,0) rectangle (.5,.5);
		
		\node[inner sep=.1] (S0) at (.5,1) {};
		\draw[fill=black]  (S0) ellipse (0.05 and 0.05);
		\draw[fill=black]  (R0) ellipse (0.05 and 0.05);
		\draw[] (.5,.5) -- (S0);
		\node at (-.1,.25) {$\idbf{}{x}{}$};
		
		\node (dPR) at (-.9,0) {};
	\end{scope}
	
	\node[inner sep=1] (A0) at (3.9,3.95) {};
	\begin{scope}[scale=1, rotate around={0:(A0)}, shift={(A0)}, color = col_teal!50!black]
		\begin{scope}[ rotate around={45:(A0)}]
			\draw[fill=col_teal!50!black]  (0,0) ellipse (0.05 and 0.05);
			\draw[draw=col_teal!50!black] (-.125,-.125) node (v1) {} rectangle ++(0.25,0.25) node (v4) {};
			\node[rotate=0] at (-.4,.0) {$A$};
		\end{scope}
	\end{scope}
	
	\node[inner sep=1] (A1) at (4.65,3.) {};
	\begin{scope}[scale=1, rotate around={0:(A1)}, shift={(A1)},color = col_teal!50!black]
		\clip  (-4,-1.7) rectangle (1.8,2);
		\begin{scope}[ rotate around={45:(A1)}]
			\draw[fill=col_teal!50!black]  (0,0) node (v2) {} ellipse (0.05 and 0.05);
			\draw[draw=col_teal!50!black] (-.125,-.125) node (v1) {} rectangle ++(0.25,0.25) node (v4) {};
		\end{scope}
		\draw[dashed]  (A1) ellipse (2.68 and 2.68);
		\draw[dotted]  (A1) ellipse (3.68 and 3.68);
		\draw[dotted]  (A1) ellipse (1.68 and 1.68);
		\node[rotate=0] at (.4,.0) {$A'$};
	\end{scope}
	
	\node[inner sep=1] (tA) at (4.6,.55) {};
	\begin{scope}[scale=1, rotate around={0:(tA)}, shift={(tA)}, col_teal!50!black]
		\begin{scope}[ rotate around={45:(tA)}]
			\draw[fill=col_teal!50!black, color = col_teal!50!black]  (0,0) node (v3) {}  ellipse (0.05 and 0.05);
	
		\end{scope}
		\node[rotate=0] at (.4,.0) {$\idbf{}{t}{A}'$};
	\end{scope}
	
	\node[col_teal!50!black] (AM) at (4.55,2.77) {};
	\draw[color=col_teal!50!black, fill=col_teal!50!black] (AM) {}  ellipse (0.05 and 0.05);
	
	\draw[]  plot[smooth, tension=.7] coordinates {(-0.5,1.75) (0.5,1.5) (3,3) (5.5,2.5) (6.55,2.75)};
	
	\draw[dashed, very thin]  (tA) ellipse (2.6 and 0.5);
	\draw[dotted]  (tA) ellipse (3.34 and 0.7);
	\draw[dotted]  (tA) ellipse (1.86 and 0.3);
	\node (dmU) at (1.28,0.54) {};
	\draw[dashdotted, very thin]  (dPR.center) -- (dmU.center);
	\draw[dashdotted, very thin]  (R0.center) -- (t0.center);
	\draw[latex-latex, color = orange]  (t0.center) -- (dmU.center);
	\draw [dashed, very thin] (t0.center) -- (tA.center);
	
	\node[inner sep=1]  at (6.5,2.4) {$\M$};
	\node (dP) at (1.05,2.29) {};
	\draw[latex-]  (A0) -- (A1);
	\draw[dashed, very thin, color = col_sage!60!black]  (A1.center) -- (R0);
	\draw[-latex, thick,col_sage!60!black]  (R0) -- (dP.center);
	\draw[dashed, very thin, col_sage!60!black]  (dP.center) -- (dPR.center);
	\draw[-latex, thick, color = teal!60!white]  (R0.center) -- (dPR.center);
	
	\node at (4.7,3.6) {$\idbf{}{r}{RS}$};	
	\node[color = col_sage!60!black] at (.7,4) {$\id{}{R}{d}$};
	\node[color=orange] at (.75,.5) {$\id{}{R}{d\U}$};
	\node[col_sage!60!black] at (0.7,2.4) {$\idbf{}{p}{d}$};
	\node[color = teal!80!white] at (.65,1.95) {$\idbf{}{p}{dT}$};

	\draw[dashdotted, very thin, col_teal!50!black]  (AM.center) -- (v3.center);
	\draw[dashdotted, very thin, col_teal!50!black]  (AM.center) -- (A1.center);
	
	\node (Rd1) at (1.12,4) {};
	\node (Rd2) at (2.07,3.73) {};
	
	\draw[dashed, very thin, color = col_sage!60!black]  (v2) -- (Rd1.center);
	\draw[latex-latex,color = col_sage!60!black]  (Rd2.center) -- (Rd1.center);
	\node at (1.5,3.5) {$S$};
	\draw[fill=black]  (R0) ellipse (0.05 and 0.05);
	
\end{tikzpicture}
	\caption{Depiction of the proposed uncertainty projection treating the original covariance $\id{}{R}{d}$ as the vector $\idbf{}{p}{d}$, projecting it onto the tangent plane and deriving the new radial uncertainty $\id{}{R}{d\U}$ by mapping $\idbf{}{p}{dT}$ to $\U$.}
	\label{fig:proj_distance_cov}
\end{figure}
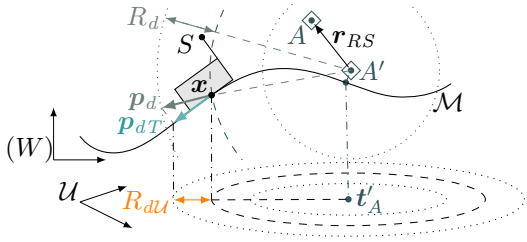

Formalizing a similar method for distance measurements analytically is significantly more complex and generally infeasible for generic terrain models.
Therefore, our approach focuses on the vicinity of the current estimate $\idbf{}{t}{R}$ as depicted in Fig. \ref{fig:proj_distance}. Assuming a reasonably consistent filter, the true position is presumed to reside within the $3\sigma$ bound of this estimate, as determined by its covariance $\idbf{}{P}{t}$. This region on $\U$ is sampled and mapped onto $\M$, generating a set of viable points in the world frame, $\left\{\idbf{}{p}{R}\right\}$. To project the range measurement $\id{}{z}{d}$ onto the surface, the corresponding sphere, centered at the static anchor $\idbf{W}{r}{A}$, is shifted by $\idbfhat{}{R}{WR}\idbf{R}{r}{RS}$ to account for the on-board sensor offset and numerically intersected with $\left\{\idbf{}{p}{R}\right\}$. This yields a set of feasible positions, $\left\{\idbf{}{p}{m}\right\}$, that inherently consider the surface geometry for the measurement.
Mapping these $N$ points back onto the chart to attain $\left\{\idbf{}{t}{m}\right\}$, and deriving an equivalent anchor position  $\idbf{}{t}{A}' \in \U$, for instance by mapping the closest point on the manifold, enables defining a new range measurement on $\U$ via the average distance to the samples $\idbf{}{t}{m,i} \in \left\{\idbf{}{t}{m}\right\}$
\begin{gather}
	\id{}{z}{d\U} = \frac{1}{N} \sum_{i}^{N}\left\|\idbf{}{t}{A}' - \idbf{}{t}{m,i}\right\|_2.
\end{gather}
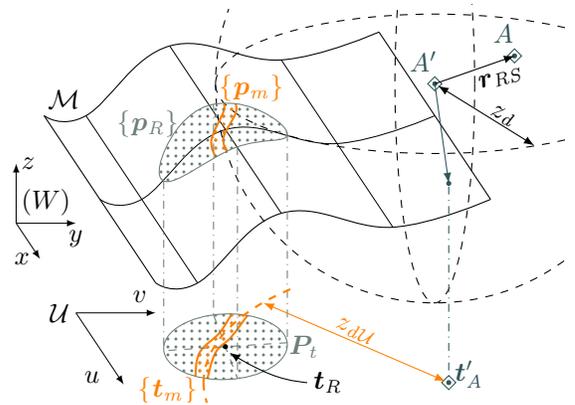
\begin{figure}[b]
	\centering
	\usetikzlibrary{patterns}
\begin{tikzpicture}[scale=.53]
	\clip  (-3.6,-3) rectangle (10.25,7.);

	\node (M) at (0,0) {};
	\begin{scope}[scale=.7, rotate around={0:(M)}, shift={(M)}]
		
		\node (M1) at (0,0) {};
		\node (M2) at (-4,6) {};
		\node (M3) at (1.5,0) {};
		\node (M4) at (-2.5,6) {};
		\node (M5) at (4.5,2.5) {};
		\node (M6) at (0.5,8.5) {};
		\node (M7) at (8.5,2) {};
		\node (M8) at (4.5,8) {};
		\node (M9) at (12,3) {};
		\node (M10) at (8,9) {};
		
		\draw  plot[smooth, tension=.7] coordinates {(0,0) (1.5,0) (4.5,2.5) (8.5,2) (12,3)};
		\begin{scope}[scale=1, shift={(-2,3)}]
			\draw[very thin]  plot[smooth, tension=.7] coordinates {(0,0) (1.5,0) (4.5,2.5) (8.5,2) (12,3)};
		\end{scope}
		\begin{scope}[scale=1, shift={(-4,6)}]
			\draw  plot[smooth, tension=.7] coordinates {(0,0) (1.5,0) (4.5,2.5) (8.5,2) (12,3)};
		\end{scope}
		\draw (M2.center) edge (M1.center);
		\draw[very thin]  (M4.center) edge (M3.center);
		\draw[very thin]  (M6.center) edge (M5.center);
		\draw[very thin]  (M8.center) edge (M7.center);
		\draw  (M10.center) edge (M9.center);
	\end{scope}

	\node (W0) at (-3.5,1.5) {};
	\begin{scope}[scale=1.5, rotate around={0:(W0)}, shift={(W0)}]
		\draw [-latex, very thin] (0,0)--(1,0);
		\draw [-latex, very thin] (0,0)--(0,1);
		\draw [-latex, very thin] (0,0) node (v2) {}--(0.4,-0.6);
		\node at (.5,.35) {$(W)$};
		\node at (.1,-.6) {$x$};
		\node at (1,-.25) {$y$};
		\node at (0.2,1) {$z$};
	\end{scope}
	
	\node (U0) at (-2,-.75) {};
	\begin{scope}[scale=1, rotate around={0:(U0)}, shift={(U0)}]
		
		\begin{scope}[scale=2, shift={((0,0)}]
			\draw [-latex,very thin] (0,0)--(1,0);
			\draw [-latex,very thin] (0,0)--(0.6,-0.9);
			\node at (-.2,0) {$\U$};
			\node at (.8,.2) {$v$};
			\node at (.2,-.8) {$u$};
		\end{scope}
		
		\draw[thick,color=orange] plot[smooth, tension=.7] coordinates {(4.3,-.06) (3.9,-.6) (3.5,-1) (3.3,-1.65)};
		\draw[thick,color=orange]  plot[smooth, tension=.7] coordinates {(3.9,-.03) (3.6,-.5) (3.1,-1) (3.0,-1.59)};

		\node (t0) at (3.75,-.85) {};
		\begin{scope}[scale=1, rotate around={0:(t0)}, shift={(t0)}]
			\node (t1) at  (1.55,.1) {};
			\node (t2) at (-1.55,-.1) {};	
			\node (t3) at  (-.3,.8) {};
			\node (t4) at (.3,-.8) {};
			
			\node (t11) at  (1.53,-.09) {};
			\node (t22) at (-1.53,.09) {};	
			\node (t33) at  (.25,.82) {};
			\node (t44) at (-.25,-.82) {};
			
			\draw[dashed, color=col_sage!60!black] (t1.center) -- (t2.center);
			\draw[dashed, color=col_sage!60!black]  (t3.center) -- (t4.center);
			\draw[pattern=dots, pattern color=col_sage!60!black, draw=col_sage!60!black ]  plot[smooth cycle, tension=1] coordinates {(t11) (t33) (t22) (t44)};
			\draw[fill=black]  (0,0) node (v3) {} ellipse (0.05 and 0.05);
		\end{scope}
		
		\draw[pattern=dots, pattern color = orange, draw=none] plot[smooth, tension=.7] coordinates {(3.9,-.03) (3.6,-.5) (3.1,-1) (3.0,-1.5) (3.3,-1.6) (3.5,-1) (3.9,-.6) (4.3,-.08)};
		
	\end{scope}

	\node[inner sep=1] (A0) at (7,5) {};
	\begin{scope}[scale=1, rotate around={0:(A0)}, shift={(A0)}]
		\begin{scope}[ rotate around={45:(A0)}]
			\draw[fill= col_teal!50!black, color = col_teal!50!black]  (0,0) ellipse (0.05 and 0.05);
			\draw[draw= col_teal!50!black] (-.125,-.125) node (v1) {} rectangle ++(0.25,0.25) node (v4) {};
			\node[rotate=0] at (.2,.6) {\textcolor{col_teal!50!black}{$A'$}};
		\end{scope}
		
		\draw[dashed] (5.5,0) arc(0:20:5.5cm and 5.5cm);
		\draw[dashed] (5.5,0) arc(0:-200:5.5cm and 5.5cm);
		\draw[dashed] (5.5,0) arc(0:-180:5.5cm and 1.75cm);
		\draw[dashed] (5.5,0) arc(0:180:5.5cm and 1.75cm);
		\draw[dashed] (1,0) arc(0:-205:1cm and 5.5cm);
		\draw[dashed] (1,0) arc(0:25:1cm and 5.5cm);
	\end{scope}
	
	\node[inner sep=1] (A1) at (9,5.7) {};
	\begin{scope}[scale=1, rotate around={0:(A1)}, shift={(A1)}, color = col_teal!50!black]
		\begin{scope}[ rotate around={45:(A1)}]
			\draw[fill= col_teal!50!black, color = col_teal!50!black]  (0,0) ellipse (0.05 and 0.05);
			\draw[draw = col_teal!50!black] (-.125,-.125) node (v1) {} rectangle ++(0.25,0.25) node (v4) {};
			\node[rotate=0] at (.2,.6) {\textcolor{col_teal!50!black}{$A$}};
		\end{scope}
	\end{scope}
	
	\draw[thick,color = orange]  plot[smooth, tension=.7] coordinates {(1.8,3.31) (1.7,3.65) (2,4.25) (2,4.5) };
	\draw[thick,color = orange]  plot[smooth, tension=.7] coordinates {(1.7,4.5) (1.7,4.3) (1.4,3.6) (1.5,3.17) };

	\draw[pattern=dots, pattern color=col_sage!60!black, draw =col_sage!60!black]  plot[smooth cycle, tension=.8] coordinates {(3.3,3.74) (2.,4.5) (.65,4) (0.2,2.56) (2,3.4)};
	\node (e1) at (3.3,3.75) {};
	\node (e2) at (0.2,2.555) {};
	\node (e3) at (1.45,4.485) {};
	\node (e4) at (2.05,3.41) {};
	
	\draw[dashdotted, color=col_sage!60!black]  (e1.center) edge (t1.center);
	\draw[dashdotted, color=col_sage!60!black]  (e2.center) edge (t2.center);
	\draw[dashdotted, color=col_sage!60!black]  (e3.center) edge (t3.center);
	\draw[dashdotted, color=col_sage!60!black]  (e4.center) edge (t4.center);
	
	\draw[pattern=dots, pattern color = orange, draw=none]  plot[smooth cycle, tension=.5] coordinates {(1.7,4.5) (1.7,4.3) (1.4,3.6) (1.5,3.25) (1.8,3.32) (1.7,3.65) (2,4.25) (2,4.49) };
	
	\draw[thick, dashed, orange] (1.2,-2.5) arc(-0:52:-5.8cm and 3cm);
	\draw[thick, dashed, orange] (1.2,-2.5) arc(-0:-15:-5.8cm and 3cm);
	\node[inner sep=1] (tA) at (7.35,-2.5){};
	\begin{scope}[ rotate around={45:(tA)}, shift={(tA)}]
		\draw[fill=col_teal!50!black, color = col_teal!50!black]  (0,0) ellipse (0.05 and 0.05);
		\draw[draw=col_teal!50!black] (-.125,-.125) node (v1) {} rectangle ++(0.25,0.25) node (v4) {};
		\node[rotate=0] at (.5,-.2) {\textcolor{col_teal!50!black}{$\idbf{}{t}{A}'$}};
	\end{scope}
	
	\node at (-2.3,4.6) {$\M$};
	\node[inner sep=1] (AM) at (7.35,2.5) {};
	\draw[fill=col_teal!50!black, color = col_teal!50!black]  (AM) ellipse (0.05 and 0.05);
	\node (dm) at (9.5,3.45) {};
	\draw[latex-latex] (A0) -- (dm.center);
	\node[rotate=-24] at (8.7,4.2) {$\id{}{z}{d}$};
	\draw[-latex] (A0.center) -- (A1.center);
	\draw[-latex , color = col_teal!50!black] (A0.center) -- (AM.center);
	\node[rotate=17] at (8.65,5.15) {$\idbf{}{r}{RS}$};
	\draw[dashdotted, color = col_teal!50!black] (AM) -- (tA);
	\node (dtm) at (2.7,-0.5) {};
	\draw[latex-latex, color = orange] (tA) -- (dtm.center);
	\node[rotate=-20] at (5,-1.2) {\textcolor{orange}{$\id{}{z}{d\U}$}};
	\node at (3.7,-1.6) {\textcolor{col_sage!60!black}{$\idbf{}{P}{t}$}};
	\node[color=orange] at (0.3,-2.7) {$\left\{\idbf{}{t}{m}\right\}$};
	\node at (-.2,4) {\textcolor{col_sage!60!black}{$\left\{\idbf{}{p}{R}\right\}$}};
	\node[color=orange] at (2.4,4.9) {$\left\{\idbf{}{p}{m}\right\}$};
	\draw[-latex]  plot[smooth, tension=.9] coordinates {(3.8,-2.5) (2.9,-2.3) (1.85,-1.75)};
	\node at (4.3,-2.5) {$\idbf{}{t}{R}$};
\end{tikzpicture}
	\caption{Illustration of the projected distance measurement $\id{}{z}{d\U}$ on the chart with the sampled $3\sigma$ region around the estimate $\idbf{}{t}{R}$, where the points projected to the manifold $\left\{\idbf{}{p}{R}\right\}$ are intersected with the range data $\id{}{z}{d}$ of the shifted anchor $A'$ to generate a set of viable positions $\left\{\idbf{}{t}{m}\right\}$ on $\U$.}
	\label{fig:proj_distance}
\end{figure}
Modeling the uncertainty of the newly formed measurement, the radial covariance $\id{}{R}{d}$ of the original data is projected towards the on-board sensor. This direction corresponds to the unit vector $\idbf{W}{n}{AS}$, pointing from the anchor shifted by 
$\idbf{W}{r}{RS}$ to the current position estimate. Trivially, this offset introduces additional uncertainty based on the state covariance $\idbf{}{P}{x}$. The resulting directional uncertainty vector $\idbf{W}{p}{d}$ is thus augmented via the Jacobian of the shift $\idbf{}{J}{d}$ w.r.t. $\idbf{}{x}{}$. To account for the constraint that uncertainties occur only along $\M$, the surface normal component of $\idbf{W}{p}{d}$ is removed before mapping the vector $\idbf{W}{p}{dT}$ to 
$\U$, as depicted in Fig. \ref{fig:proj_distance_cov}. The radial covariance on the chart, $\id{}{R}{d\U}$, is then given by
\begin{gather}
	\idbf{W}{p}{d} = \idbf{W}{n}{AS}\id{}{R}{d} + \idbf{}{J}{d}\idbf{}{P}{x}\idbf{}{J}{d}^T
	\\
	\id{}{R}{d\U} = \left\|\pfpx{\sigma}{\bb{p}} \idbf{}{p}{dT} \right\|_2.
\end{gather}

\section{EXPERIMENTS}
To validate the proposed strategies, we simulate a vehicle equipped with an odometry system, one global pose sensor and a range measurement unit paired with two static anchors, mimicking configurations typical in industrial UGVs. Bivariate b-spline surfaces are utilized to generate relevant scenarios, as they can effectively approximate diverse terrain types. Defined via spline basis functions $N_{i}(x), N_j(y)$ and control points $\idbf{}{P}{i,j}$, these differentiable explicit surfaces simplify the definition of a chart map $\sigma$ and its inverse $\sigma^{-1}$
\begin{gather}
	S(x,y) = \sum_{i=0}^{n}\sum_{j=0}^{m}\idbf{}{P}{i,j}N_{i}(x)N_j(y) 
	\\
	\sigma(\idbf{}{p}{} \in \M) = \sigma(x,y,z) = \begin{bmatrix} u \\ v \end{bmatrix} = \begin{bmatrix} x \\ y \end{bmatrix} \in \U
	\\
	\sigma^{-1}(\idbf{}{t}{} \in \U) = \sigma^{-1}(u,v) = \begin{bmatrix} x \\ y \\ z \end{bmatrix} = \begin{bmatrix} u \\ v \\ S(u,v) \end{bmatrix} \in \M.
\end{gather}

\subsection{Evaluation Method}
Integrated into the validated \textit{MaRS} framework \cite{brommer2021}, two M-ESEKFs, one employing the measurement models in $\RIII$ (M-ESEKF) and the other the projected corrections (MP-ESEKF), were tested. Their performance was benchmarked against a classical, constrained ESEKF (C-ESEKF), which fixes elevation, roll, and pitch via geometry-based pseudo-measurements. Multiple Monte Carlo simulations across various terrains were conducted with $N=100$ trials, generating truly zero-mean Gaussian-distributed measurements from ground truth trajectories, while dynamically enabling and disabling individual sensors. At each time step, the \textit{Root Mean Square Error} (RMSE) of position and orientation, as well as the \textit{Average Normalized Estimator Error Squared} (ANEES) were computed and averaged over all $N$ trials
\begin{gather}
	RMSE_k = \sqrt{\frac{1}{N}\sum_{i=1}^{N}\idbf{}{e}{i_k}^2}
	\\
	ANEES_k = \frac{1}{Nm}\sum_{i=1}^{N}\idbf{}{e}{i_k}^T\idbf{}{P}{i_k}^{-1}\idbf{}{e}{i_k}.
\end{gather}
Here, $\idbf{}{e}{i_k}$ denotes the estimator error with $m$ elements and $\idbf{}{P}{i_k}$ represents the corresponding covariance for the $i^\text{th}$ run at sample step $k$. The RMSE provides a metric for the filter's average accuracy, while the ANEES serves as an indicator for its consistency. An estimator achieving an ANEES value close to $1$ is deemed more credible and thus consistent \cite{li2012}.

\subsection{Results}
\begin{figure}[b]
	\centering
	\includegraphics[scale=.65, trim= 5ex 19ex 5ex 16ex, clip]{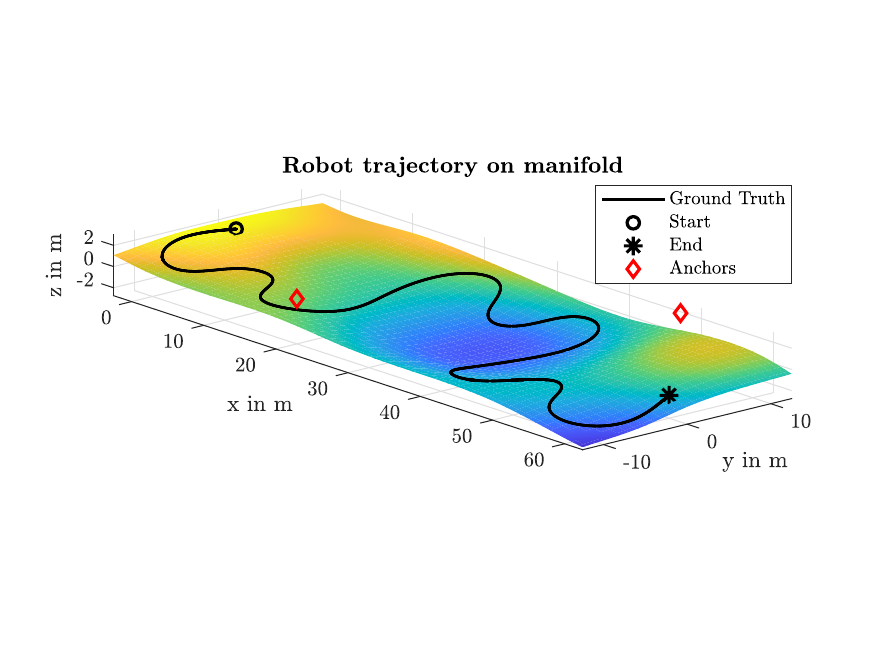}
	\caption{The exemplary evaluation scenario with the ground truth trajectory in black and the two static range measurement anchors in red.}
	\label{fig:trajectory}
\end{figure}

\begin{figure}[t]
	\centering
		\vspace{1.2ex}
	\includegraphics[scale=.67, trim= 0ex 1.1ex 5ex .9ex, clip]{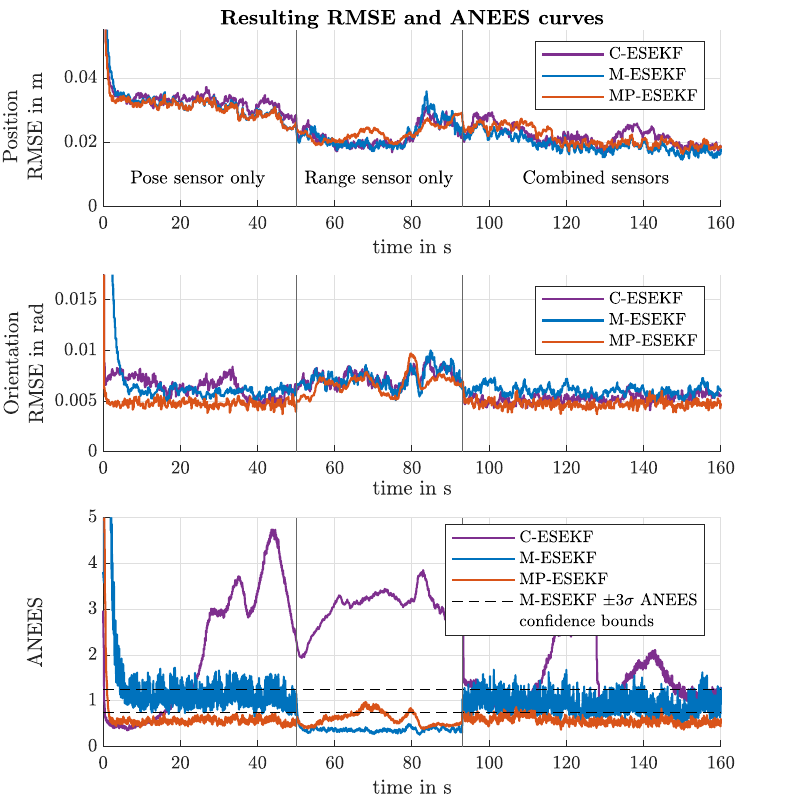}
	\caption{The resulting RMSE and ANEES plots of the compared estimators on the selected example scenario averaged over $100$ trials. The vertical lines indicating the switch from utilizing only pose measurements to solely relying on range data and finally combining both inputs.}
	\label{fig:plots}
\end{figure}

Fig. \ref{fig:trajectory} illustrates the trajectories from one of the simulated scenarios, representative of the overall validation results, while Fig. \ref{fig:plots} presents the corresponding performance metrics and Tab. \ref{tab:sensors} summarizes the sensor configurations. Note that orientation measurements were generated as Tait–Bryan angles and converted to quaternions to match our formulation.

Although the M-ESEKF exhibits slower initial convergence of rotational errors compared to other methods and the MP-ESEKF slightly outperforms in tracking the orientation, overall differences in RMSE, and therefore accuracy, are minimal across the analyzed estimators. All implementations enhance localization precision relative to the raw measurement uncertainties, achieving a positional RMSE below \SI{0.038}{\meter} and a maximum orientation error of \SI{0.01}{\radian}. When combining all available inputs, these errors stabilize at approximately \SI{0.02}{\meter} and \SI{0.005}{\radian}, respectively. 

While this suggests that all filters are comparably precise, their consistency differs markedly. The C-ESEKF tends to produce overconfident estimates, with ANEES values fluctuating considerably throughout the trials. By contrast, the manifold-based filters exhibit higher consistency, especially during the initial and final trajectory segments. The M-ESEKF with analytical measurement models is more credible on average, with a mean ANEES closer to $1$. The projected corrections of the MP-ESEKF however stabilize the estimator consistency by significantly reducing ANEES noise, mostly confining it to within the \SI{99}{\percent} confidence bounds, albeit introducing slight under-confidence. These more conservative estimates likely result from the numerical approaches taken and could potentially be improved by developing analytical formulations, more accurately representing the true uncertainties.

\begin{table}[h]
	\centering
	\caption{Simulated Sensor Configuration}
	\label{tab:sensors}
	\renewcommand{\arraystretch}{1.25}
	\begin{tabular}{lcc}\hline
		\textbf{Sensor} & \textbf{Rate} & \textbf{Measurement Standard Deviation} \\ \hline
		Odometry & \SI{20}{\Hz} & Linear: \SI{0.02}{\meter/\second}, Angular: \SI{0.01}{\radian/\second} \\ 
		Pose & \SI{5}{\Hz} & Position: \SI{0.03}{\meter}, Orientation: \SI{0.01}{\radian} \\ 
		Range & \SI{10}{\Hz} & Distance: \SI{0.05}{\meter} \\\hline 
	\end{tabular}
\end{table}

Notably, achieving competitive C-ESEKF results required heavy manual tuning of the pseudo-measurements to specific terrains and sensor setups, balancing RMSE performance with estimator consistency as measured by the ANEES. Generic settings resulted in either strong overconfidence with an ANEES exceeding $50$ or diverging estimates. The M-ESEKF requires no tuning, and the MP-ESEKF’s sampling settings can be generalized from sensor specifications.

In terms of computational costs, the closest-point projections to the manifold imposed minimal overhead in our tests. Range-region sampling was however noticeably slower, due to frequent b-spline evaluations within $\sigma^{-1}$. The effort varies with state, uncertainties and surface model, hence exact complexity estimates are difficult. Without dedicated optimization, the projected range measurements were processed in under $\SI{6}{\milli\second}$ as opposed to $\SI{20}{\micro\second}$ of the analytical ones, which is generally still acceptable for common applications. 

\section{CONCLUSIONS AND OUTLOOK}
The proposed filter formulation effectively improves consistency of EKF-based localization for terrestrial vehicles. By combining a manifold state with a locally planar odometry model, the filter achieves accurate, consistent results, remaining nearly invariant to the surface geometry and requiring no manual tuning. The introduction of novel measurement projection techniques further stabilizes the estimator's performance, resulting in a more predictable filter with reliable credibility for multi-sensor pose estimation.

Simulated experiments with relevant sensor classes demonstrate that the proposed methods outperform classical approaches in scenarios reflective of real-world applications. Consequently, the M-ESEKF and its extension, the MP-ESEKF, present a viable and practical solution for achieving consistent and accurate pose estimation in UGVs.

A key limitation of the M-ESEKF is the need for surface smoothness and accurate geometry models. While such data may be available in structured indoor sites or for repeated inspection tasks, future work could ease this dependency by estimating terrain layout online and investigating the impact of manifold errors. Furthermore, real-world experiments could provide valuable insights on the filter's performance.

\bibliographystyle{IEEEtran}
\bibliography{IEEEabrv, bibliography}

\end{document}